\documentclass[11pt,a4paper]{article}
\usepackage[hyperref]{acl2020}
\usepackage{times}
\usepackage{latexsym}

\usepackage{microtype}

\aclfinalcopy 

\title{Are Classes Clusters?}

\author{
    Kees Varekamp \\
    Stanford Center for Professional Development \\
    \texttt{kees.varekamp@gmail.com}
}

\date{}

\begin{document}
\maketitle
\begin{abstract}
Sentence embedding models aim to provide general purpose embeddings for sentences. Most of the models studied in this paper claim to perform well on STS tasks – but they do not report on their suitability for clustering.

This paper\footnote{Project was completed as part of the XCS224U professional course} looks at four recent sentence embedding models (Universal Sentence Encoder \citep{cer-etal-2018-universal}, Sentence-BERT \citep{reimers2019sentencebert}, LASER \citep{artetxe-schwenk-2019-massively}, and DeCLUTR \citep{giorgi2020declutr}). It gives a brief overview of the ideas behind their implementations.

It then investigates how well topic classes in two text classification datasets (Amazon Reviews \citep{ni-etal-2019-justifying} and News Category Dataset \citep{news-dataset}) map to clusters in their corresponding sentence embedding space. While the performance of the resulting classification model is far from perfect, it is better than random. 

This is interesting because the classification model has been constructed in an unsupervised way. The topic classes in these real life topic classification datasets can be partly reconstructed by clustering the corresponding sentence embeddings.

\end{abstract}

\begin{table*}
\centering
\begin{tabular}{lrr}
\hline
\textbf{Model} & \textbf{Cluster F1} & \textbf{LogReg F1} \\
\hline
Sentence-BERT   &        0.22 &      0.53 \\
Universal Sentence Encoder     &        \textbf{0.29} &      \textbf{0.56} \\
DeCLUTR &        0.22 &      0.56 \\
LASER   &        0.10 &      0.42 \\
TfidfVectorizer     &        0.07 &      0.55 \\
\hline
\end{tabular}
\caption{Average F1 scores of unsupervised cluster classifier and logistic regression classifier. Averages taken over two datasets: Amazon Reviews dataset and News dataset. Bold is best of column.}
\label{table:mainresults}
\end{table*}

\section{Introduction}

Since the success of universal word embeddings such as word2vec \citep{mikolov2013} and GloVE \citep{pennington-etal-2014-glove} the interest in the sentence equivalent of such embeddings has been increasing.

Sentence embeddings promise to be useful for many tasks in natural language processing. They can provide standardized inputs for custom models such as topic classifiers or sentiment analysis models. But they are even more useful for tasks that involve computing the semantic similarity between many sentences.

It is perfectly possible to train a well-performing model on sentence similarity without any pre-computed sentence embeddings. But as was pointed out by \citep{reimers2019sentencebert}, you will need to run one inference for every similarity between two sentences. For a similarity matrix between $n$ sentences you will need to run $n(n-1)/2$ inferences. 

And when the model is large (a Transformer based model for example), these costs add up. The big advantage of universal sentence embedding models is that the vector representations of the sentences can be pre-computed individually. If the proximity of two vectors in the embedding space closely matches the semantic similarity of the two corresponding sentences (and most sentence embedding models indeed aim for this), the similarity calculation becomes as simple as the opposite of an L2 difference or a cosine similarity. Only $n$ potentially expensive embeddings need to be pre-computed instead of $n(n-1)/2$, after which the above similarity matrix can be produced much more efficiently.

This opens the door to many useful applications that involve semantic search in one way or another,  two of the most prominent ones being k-nearest-neighbour search and cluster analysis.

This paper focuses on cluster analysis. It will try to answer the following question:  Do topic classes in classification datasets correspond to clusters in the corresponding embedding space? 

It attempts to answer this question in a very straightforward manner: by creating a classification model based on an unsupervised clustering algorithm. We use the usual metrics to evaluate the performance of this unsupervised model. And finally we compare the result to a logistic regression model which acts as the standard supervised baseline.

It turns out (Table \ref{table:mainresults}) that it is in fact somewhat possible to use clusters in the embedding space of a topic classification corpus to classify the corresponding topics. The approach is nowhere near state of the art, but the fact that it works is interesting and may be useful for unsupervised text classification applications.

\section{Related work: SentEval}
The canonical benchmark for sentence embedding models is SentEval \citep{conneau2018senteval}. SentEval consists of 2 parts: The first part is a series of benchmarks that are called Downstream tasks - these are tasks that use the embeddings to perform a series of common NLP benchmark tasks. The second part is called Probing Tasks. These are a series of tasks that attempt to reconstruct certain linguistic properties of the original sentence.

The tasks report a variety of scores, but all of them include at least either accuracy or Spearman's rank correlation coefficient.

The models in this study have been evaluated on SentEval for comparison with the Cluster Classifier experiment. Results are in Table \ref{table:sentevalresults}.

\section{Data}
\subsection{Amazon Reviews Dataset}
The first dataset that we investigate is the Amazon Reviews Dataset \citep{ni-etal-2019-justifying}. The data was downloaded from \url{https://nijianmo.github.io/amazon/index.html} and was sampled down to contain a 1000 examples for each category. There are 29 categories of products in the dataset (for example "Books", "Appliances", "Toys and Games", etc). 

There are titles in the reviews but they are not always equally suitable for topic classification: They tend to be short and generic ("Great Product"). So rather than using the titles we pick the first sentence in the review that is longer than 10 words and shorter than 20 words.

This approach aims to select sentences that are long enough to be representative of the topic label, and filter out short sentences that may be too generic (such as ‘I really like this product’). But it also avoids selecting sentences that are too long for the sentence embedding models to be able to accurately represent them (the representations will get too diluted). \\

The approach gives pretty good results for finding meaningful examples. Some examples:\\
\textit{(Luxury and Beauty)}: "It doesnt stay on or protect your hands through washing." \\
\textit{(Fashion)}:"Bought the just as gorgeous rich, red Fireman's scarf, too."\\
\textit{(Magazine Subscriptions)}: "I enjoy this magazine very much  And especially this one on COLOR."\\
\textit{(Automotive)}: "Great for routine maintenance tasks such as oil changes, etc."\\

\subsection{News Dataset}
The second dataset collects news from the Huffington Post \citep{news-dataset} and was downloaded from \url{https://www.kaggle.com/rmisra/news-category-dataset}. It has also been sampled down to contain a 1000 examples of each news category. There are 40 news categories, for example "SPORTS", "CRIME", etc.

Each example contains metadata, a short description, a headline and a category. We choose the headline for our analysis. \\

Some examples:\\
\textit{(CRIME)}: "There Were 2 Mass Shootings In Texas Last Week, But Only 1 On TV"\\
\textit{(POLITICS)}: "How The Chinese Exclusion Act Can Help Us Understand Immigration Politics Today"\\
\textit{(WOMEN)}: "The 20 Funniest Tweets From Women This Week"\\
\textit{(TECH)}: "Self-Driving Uber In Fatal Accident Had 6 Seconds To React Before Crash"\\

\section{Models}
\subsection{Embedding Models}
This section describes the universal sentence embedding models that we will investigate.

\subsubsection{Predecessors}
A lot of work has been done in this space, and there is not enough room here to go over all of it. So this summary only includes four of the most recent models in some detail (USE, SBERT, LASER, and DeCLUTR). It is omitting important ground work, in particular the following 3 studies and their resulting models:
\begin{itemize}
    \item SkipThought \citep{kiros2015skipthought}: Basically “Thou shalt know a word by the company it keeps”, but extended to sentences. An unsupervised method based on the idea that sentences that are close to each other are more similar than sentences that are far away from each other, much like word2vec and glove. 
    \item InferSent \citep{conneau-etal-2017-supervised}: max pooling on the hidden states of a biLSTM trained on the Stanford Natural Language Inference dataset
    \item QuickThoughts \citep{logeswaran2018efficient}: A non-generative version of SkipThought – rather than trying to reconstruct neighbouring sentences, it tries to train a classifier for predicting neighbouring/not neighbouring.
\end{itemize}

\subsubsection{Universal Sentence Encoder}
Universal Sentence Encoder (USE) \citep{cer-etal-2018-universal} is a model developed by Google. The goal of USE is to provide easy-to-use sentence embeddings with good transfer performance.

There are actually 2 implementations of the encoder:
\begin{itemize}
    \item A Transformer-based model using (self) attention that mean-pools the word embeddings to produce sentence embeddings
    \item A Deep Averaging Network (DAN) that simply averages word embeddings and bi-gram embeddings and feeds the resulting representation into a straightforward deep neural network. The DAN trades accuracy for performance, with respect to the transformer, which scales with the square of the sentence length.
\end{itemize}

The models are trained on a set of self-supervised tasks and on SNLI 

The unsupervised tasks are trained on Wikipedia, web news, web question-answer pages and discussion forums

One interesting note in the paper mentions that for STS tasks the authors use the angular distance (the arccos of the cosine similarity) as a distance measure, because “arccos converts cosine similarity into an angular distance that obeys the triangle inequality. We find that angular distance performs better on STS than cosine similarity.”

\subsubsection{Sentence-BERT}

Sentence-BERT (SBERT) \citep{reimers2019sentencebert} is a BERT based model that uses a Siamese network approach to fine-tune BERT on a set of tasks in order to produce useful sentence embeddings. 

For SBERT the goal is to produce embeddings that are primarily useful for computationally and/or combinatorially expensive tasks like clustering and nearest neighbour search – producing vectors that can be used in optimized computation libraries such as numpy or sklearn.

SBERT promises semantically meaningful sentence embeddings (similar sentences are close in vector space). 

It tries to accomplish this by pooling the token embeddings while training on STS tasks but using fixed weights between 2 BERTs, so that the resulting embeddings for the two input sentences are created using identical weights. It claims to be better on STS or SentEval than InferSent or USE.

SBERT trains using 3 objectives:
\begin{itemize}
    \item Classification, by pooling the embeddings of $u$ and $v$ as $[u; v ; u-v]$ as input to a softmax, using cross entropy loss
    \item Regression (presumably on STS), by using cosim(u,v) using MSE as the loss
    \item Triplet loss: Given 3 sentences; a (anchor), p (positive example) and n (negative example), train the model such that $|a-p| < |a-n|$, or in other words: minimize $|a-p| - |a-n|$.
\end{itemize}
SBERT is trained on SNLI and MultiNLI. It does well on STS tasks (as expected)

\subsubsection{LASER}
LASER \citep{artetxe-schwenk-2019-massively} is a bit different from the other models in this summary in that it is primarily interested in universal language agnostic sentence embeddings. It is trained on a large dataset consisting of 93 languages. It uses a relative simple biLSTM to encode the input text – BPE tokenized text of any language is fed into the same encoder. This creates a fixed size language agnostic embedding that is then used in the decoder to translate the output into a specified language. The paper claims that fixed length representations are more versatile and compatible than variable length representations: “For instance, there is not always a one-to-one correspondence among words in different languages (e.g. a single word of a morphologically complex language might correspond to several words of a morphologically simple language), so having a separate vector for each word might not transfer as well across languages.” 
\begin{itemize}
    \item The embeddings are results of max pooling the hidden states of layers of the biLSTM of dimensionality 512, concatenating forward and backward hidden state into sentence representations of dimensionality 1024
    \item The authors claim that the model does well on
        \begin{itemize}
            \item XNLI (entailment)
            \item MLDoc (document classification)
            \item BUCC (finding the same sentence in another language)
            \item Tatoeba: a new cross language similarity search benchmark
        \end{itemize}
    \item In the future, the authors would like to improve the result by using a self attention encoder, pre trained word embeddings, and back translation.
\end{itemize}

\subsubsection{DeCLUTR}
DeCLUTR \citep{giorgi2020declutr} aims to produce useful universal sentence embeddings by unsupervised learning only. It points out that the best results so far have been obtained by methods that used at least some supervised learning, but that it is important to close the gap between supervised and unsupervised methods for languages and domains for which no supervised data exists.
\begin{itemize}
    \item Wants to be good at wide variety of tasks
    \item Most universal sentence embedders train supervised on SNLI or multiNLI (entailment, contradiction, neutral). Examples are InferSent, USE, and SBERT.
    \item The authors describe Skip-thought as an unsupervised generative model that uses sentence embeddings to predict words in neighbouring sentences. They mention that the generative nature of the model makes it expensive and surface focused. QuickThoughts tries to improve on this by classifying context sentences from non-context sentences rather than generating them. 
    \item The authors describe DeCLUTR’s approach as similar to SBERT, but self-supervised, and the objective as similar to QuickThoughts, but using segments rather than whole sentences. 
    \item Like SBERT’s triplet approach, it uses contrastive loss: It tries to minimize $|a-p| - |a-n|$. 
    \item Like QuickThoughts, it classifies sentences (or rather sentence segments) as near by or far away.
\end{itemize}

\subsubsection{TfidfVectorizer}
This is simply the TfidfVectorizer model from scikit-learn (\url{https://scikit-learn.org/stable/modules/generated/sklearn.feature_extraction.text.TfidfVectorizer.html}). It is a WordVectorizer (bag of words) with a TF-IDF weighting applied to it. It has been included in the experiments as a baseline for the encoder models.

\subsection{Classifiers}
This section describes the models we will be using for our experiment

\subsubsection{Cluster Classifier}
We test the main hypothesis by using a simple model:
\begin{itemize}
    \item For every dataset, we cluster the data into as many clusters as there are classes. We use k-means clustering.
    \item For every cluster, we allocate the most frequently occurring class within the cluster as its designated class
    \item This is the model. Prediction is simply a mapping from the predicted cluster of an example to the designated class of that cluster
    \item F1 (macro averaged) will give a measure of how well the model performs
\end{itemize}

\subsubsection{Logistic Regression}
For establishing a baseline on the classifier models, we use logistic regression on the embedding data to train a linear model.

\section{Experiments}
\subsection{Cluster Classifier}
The Cluster Classifier was able to retrieve some of the topic classes. Table \ref{table:mainresults} shows that the F1 scores of the Cluster Classifier are not great, but they are also not random.

Most of the classes ended up being mapped to by one (or more!) of the clusters. The cluster classifier also mixed up some classes and missed out on some other ones altogether.

The full ranking for Cluster Classifier is:

\begin{enumerate}
    \item USE
    \item DeCLUTR
    \item SBERT
    \item LASER
    \item TfidfVectorizer
\end{enumerate}

\subsection{Logistic Regression Classifier}
The Logistic Regression Classifier did significantly better than the Cluster Classifier. This is not surprising as it is a supervised model that has been given the classes in advance.

It is interesting to note that TfidfVectorizer performed reasonably well (better than SBERT) using this classifier. The full ranking for Logistic Regression is:

\begin{enumerate}
    \item USE
    \item DeCLUTR
    \item TfidfVectorizer
    \item SBERT
    \item LASER
\end{enumerate}

\subsubsection{SentEval}
If we count the number of tasks on SentEval for which each model has the highest score (see Table \ref{table:sentevalresults}, we end up with the following list:
\begin{enumerate}
    \item DeCLUTR (10 wins)
    \item SBERT (8 wins)
    \item LASER (6 wins)
    \item USE (2 wins)
\end{enumerate}

\section{Analysis}
The embedding models that were trained with STS tasks in mind (SBERT, USE, and DeCLUTR) scored better than the ones that didn't (LASER and TfidfVectorizer). This was to be expected.

But there are also unexpected results. Universal Sentence Encoder performs the worst on SentEval, yet it manages to obtain by far the best scores in both Cluster Classifier and Logistic Regression. It is unclear where this discrepancy comes from. Perhaps it is related to the fact that USE uses angular distance rather than cosine similarity. But it is difficult to dig deeper as the paper is light on details about the training data and procedures. At the very least we can conclude from this that great performance on STS tasks does not automatically lead to great performance on clustering tasks.

Over all it is encouraging that Cluster Classifier even partially works. The datasets used in this experiment are real life datasets. The topic classes used in them are not created artificially with academic purposes in mind, but have evolved out of practical considerations in real world scenario's. Yet the various sentence embedding models have managed to find a partial overlap.

\section{Conclusion}
Sentence embeddings can be clustered. In real life datasets, the resulting clusters form at least some overlap with topic classes.

This is interesting because clustering is an unsupervised analysis technique. It means that sentence embedding clusters can be used for setting up unsupervised text classification, which is a major task in real world applications such as for example customer feedback analysis.

\bibliography{anthology,acl2020}

\begin{thebibliography}{12}
\expandafter\ifx\csname natexlab\endcsname\relax\def\natexlab#1{#1}\fi

\bibitem[{Artetxe and Schwenk(2019)}]{artetxe-schwenk-2019-massively}
Mikel Artetxe and Holger Schwenk. 2019.
\newblock \href {https://doi.org/10.1162/tacl_a_00288} {Massively multilingual
  sentence embeddings for zero-shot cross-lingual transfer and beyond}.
\newblock \emph{Transactions of the Association for Computational Linguistics},
  7:597--610.

\bibitem[{Cer et~al.(2018)Cer, Yang, Kong, Hua, Limtiaco, St.~John, Constant,
  Guajardo-Cespedes, Yuan, Tar, Strope, and Kurzweil}]{cer-etal-2018-universal}
Daniel Cer, Yinfei Yang, Sheng-yi Kong, Nan Hua, Nicole Limtiaco, Rhomni
  St.~John, Noah Constant, Mario Guajardo-Cespedes, Steve Yuan, Chris Tar,
  Brian Strope, and Ray Kurzweil. 2018.
\newblock \href {https://doi.org/10.18653/v1/D18-2029} {Universal sentence
  encoder for {E}nglish}.
\newblock In \emph{Proceedings of the 2018 Conference on Empirical Methods in
  Natural Language Processing: System Demonstrations}, pages 169--174,
  Brussels, Belgium. Association for Computational Linguistics.

\bibitem[{Conneau and Kiela(2018)}]{conneau2018senteval}
Alexis Conneau and Douwe Kiela. 2018.
\newblock \href {http://arxiv.org/abs/1803.05449} {Senteval: An evaluation
  toolkit for universal sentence representations}.

\bibitem[{Conneau et~al.(2017)Conneau, Kiela, Schwenk, Barrault, and
  Bordes}]{conneau-etal-2017-supervised}
Alexis Conneau, Douwe Kiela, Holger Schwenk, Lo{\"\i}c Barrault, and Antoine
  Bordes. 2017.
\newblock \href {https://doi.org/10.18653/v1/D17-1070} {Supervised learning of
  universal sentence representations from natural language inference data}.
\newblock In \emph{Proceedings of the 2017 Conference on Empirical Methods in
  Natural Language Processing}, pages 670--680, Copenhagen, Denmark.
  Association for Computational Linguistics.

\bibitem[{Giorgi et~al.(2020)Giorgi, Nitski, Bader, and
  Wang}]{giorgi2020declutr}
John~M. Giorgi, Osvald Nitski, Gary~D. Bader, and Bo~Wang. 2020.
\newblock \href {http://arxiv.org/abs/2006.03659} {Declutr: Deep contrastive
  learning for unsupervised textual representations}.

\bibitem[{Kiros et~al.(2015)Kiros, Zhu, Salakhutdinov, Zemel, Torralba,
  Urtasun, and Fidler}]{kiros2015skipthought}
Ryan Kiros, Yukun Zhu, Ruslan Salakhutdinov, Richard~S. Zemel, Antonio
  Torralba, Raquel Urtasun, and Sanja Fidler. 2015.
\newblock \href {http://arxiv.org/abs/1506.06726} {Skip-thought vectors}.

\bibitem[{Logeswaran and Lee(2018)}]{logeswaran2018efficient}
Lajanugen Logeswaran and Honglak Lee. 2018.
\newblock \href {http://arxiv.org/abs/1803.02893} {An efficient framework for
  learning sentence representations}.

\bibitem[{Mikolov et~al.(2013)Mikolov, Sutskever, Chen, Corrado, and
  Dean}]{mikolov2013}
Tomas Mikolov, Ilya Sutskever, Kai Chen, Greg~S Corrado, and Jeff Dean. 2013.
\newblock Distributed representations of words and phrases and their
  compositionality.
\newblock \emph{Advances in neural information processing systems (NIPS)},
  pages 3111–--3119.

\bibitem[{Misra(2018)}]{news-dataset}
Rishabh Misra. 2018.
\newblock \href {https://doi.org/10.13140/RG.2.2.20331.18729} {News category
  dataset}.

\bibitem[{Ni et~al.(2019)Ni, Li, and McAuley}]{ni-etal-2019-justifying}
Jianmo Ni, Jiacheng Li, and Julian McAuley. 2019.
\newblock \href {https://doi.org/10.18653/v1/D19-1018} {Justifying
  recommendations using distantly-labeled reviews and fine-grained aspects}.
\newblock In \emph{Proceedings of the 2019 Conference on Empirical Methods in
  Natural Language Processing and the 9th International Joint Conference on
  Natural Language Processing (EMNLP-IJCNLP)}, pages 188--197, Hong Kong,
  China. Association for Computational Linguistics.

\bibitem[{Pennington et~al.(2014)Pennington, Socher, and
  Manning}]{pennington-etal-2014-glove}
Jeffrey Pennington, Richard Socher, and Christopher Manning. 2014.
\newblock \href {https://doi.org/10.3115/v1/D14-1162} {{G}love: Global vectors
  for word representation}.
\newblock In \emph{Proceedings of the 2014 Conference on Empirical Methods in
  Natural Language Processing ({EMNLP})}, pages 1532--1543, Doha, Qatar.
  Association for Computational Linguistics.

\bibitem[{Reimers and Gurevych(2019)}]{reimers2019sentencebert}
Nils Reimers and Iryna Gurevych. 2019.
\newblock \href {http://arxiv.org/abs/1908.10084} {Sentence-bert: Sentence
  embeddings using siamese bert-networks}.

\end{thebibliography}
\bibliographystyle{acl_natbib}

\appendix

\section{Supplemental Material}
\label{sec:supplemental}
\begin{table*}
\centering
\begin{tabular}{llrrrr}
\hline
{} &   \textbf{Result type} &  \textbf{SBERT} &  \textbf{DeCLUTR} &     \textbf{USE} &   \textbf{LASER} \\
\hline
STS12                 &  spearman &   \textbf{0.695} &    0.635 &   0.656 &   0.623 \\
STS13                 &  spearman &   \textbf{0.737} &    0.726 &   0.680 &   0.516 \\
STS14                 &  spearman &   \textbf{0.763} &    0.717 &   0.715 &   0.670 \\
STS15                 &  spearman &   \textbf{0.821} &    0.799 &   0.808 &   0.754 \\
STS16                 &  spearman &   \textbf{0.807} &    0.796 &   0.787 &   0.723 \\
MR                    &  accuracy &  79.830 &   \textbf{84.990} &  75.150 &  74.080 \\
CR                    &  accuracy &  86.780 &   \textbf{90.010} &  81.780 &  80.530 \\
MPQA                  &  accuracy &  86.630 &   \textbf{88.330} &  87.150 &  88.210 \\
SUBJ                  &  accuracy &  92.680 &   \textbf{95.240} &  91.800 &  91.480 \\
SST2                  &  accuracy &  84.130 &   \textbf{89.840} &  80.510 &  79.850 \\
SST5                  &  accuracy &  45.930 &   \textbf{48.550} &  42.810 &  44.250 \\
TREC                  &  accuracy &  89.200 &   91.800 &  \textbf{92.200} &  89.200 \\
MRPC                  &  accuracy &  73.860 &   74.030 &  69.620 &  \textbf{75.190} \\
SICKEntailment        &  accuracy &  81.290 &   82.100 &  \textbf{82.360} &  80.980 \\
SICKRelatedness       &  spearman &   \textbf{0.802} &    0.786 &   0.789 &   0.790 \\
STSBenchmark          &  spearman &   \textbf{0.819} &    0.794 &   0.791 &   0.778 \\
Length                &  accuracy &  64.430 &   \textbf{82.360} &  64.910 &  79.660 \\
WordContent           &  accuracy &  \textbf{73.970} &   62.150 &  70.130 &  40.380 \\
Depth                 &  accuracy &  31.440 &   34.430 &  27.430 &  \textbf{39.390} \\
TopConstituents       &  accuracy &  71.800 &   71.050 &  62.840 &  \textbf{78.490} \\
BigramShift           &  accuracy &  76.380 &   \textbf{87.630} &  59.930 &  67.590 \\
Tense                 &  accuracy &  87.000 &   \textbf{88.400} &  80.080 &  87.290 \\
SubjNumber            &  accuracy &  83.320 &   85.800 &  74.550 &  \textbf{90.220} \\
ObjNumber             &  accuracy &  81.810 &   83.130 &  72.550 &  \textbf{88.740} \\
OddManOut             &  accuracy &  56.500 &   \textbf{64.210} &  54.030 &  50.790 \\
CoordinationInversion &  accuracy &  57.080 &   66.290 &  54.300 &  \textbf{67.840} \\
\hline
\end{tabular}
\caption{SentEval scores for SBERT, DeCLUTR, USE, and LASER. SBERT scores very well on STS tasks. LASER scores well on probing tasks (linguistic properties of the input). Bold is best of row. DeCLUTR scores best over all.}
\label{table:sentevalresults}
\end{table*}

\begin{table*}
\centering
\begin{tabular}{lrrrr}
\hline
{} &  Clusters F1 &  Clusters Acc &  LogReg F1 &  LogReg Acc \\
\hline
SBERT   &        0.225 &         0.263 &      0.586 &       0.586 \\
USE     &        \textbf{0.321} &         \textbf{0.345} &      \textbf{0.618} &       \textbf{0.618} \\
DeCLUTR &        0.239 &         0.276 &      0.613 &       0.612 \\
LASER   &        0.087 &         0.113 &      0.438 &       0.446 \\
TfidfVectorizer     &        0.088 &         0.107 &      0.605 &       0.604 \\
\hline
\end{tabular}
\caption{F1 macro and accuracy scores for SBERT, DeCLUTR, USE, and LASER on the Amazon Reviews dataset. Bold is best of column.}
\label{table:amaclusterresults}
\end{table*}

\begin{table*}
\centering
\begin{tabular}{lrrrr}
\hline
  &  Clusters F1 &  Clusters Acc &  LogReg F1 &  LogReg Acc \\
\hline
SBERT   &        0.212 &         0.247 &      0.472 &       0.475 \\
USE     &        \textbf{0.256} &         \textbf{0.286} &      \textbf{0.511} &       \textbf{0.515} \\
DeCLUTR &        0.206 &         0.243 &      0.506 &       0.508 \\
LASER   &        0.112 &         0.138 &      0.398 &       0.402 \\
TfidfVectorizer     &        0.058 &         0.075 &      0.486 &       0.487 \\
\hline
\end{tabular}
\caption{F1 macro and accuracy scores for SBERT, DeCLUTR, USE, and LASER on the News dataset. Bold is best (vertically)}
\label{table:newsclusterresults}
\end{table*}

\begin{itemize}
    \item Full results of the SentEval benchmark are available in Table \ref{table:sentevalresults}. 
    \item Full results of the Cluster Classifier and Logistic Regression on the Amazon Reviews dataset are available in Table \ref{table:amaclusterresults}. 
    \item Full results of the Cluster Classifier and Logistic Regression on the News dataset are available in Table \ref{table:newsclusterresults}.     
\end{itemize}

\end{document}